\pdfoutput=1

\documentclass{article}
\usepackage{ijcai20}
\usepackage{color}
\usepackage{times}
\usepackage{soul}
\usepackage{url}
\usepackage[draft]{hyperref}
\usepackage[utf8]{inputenc}
\usepackage{graphicx}
\usepackage{dsfont}
\usepackage{amsmath}
\usepackage{amsthm}
\usepackage{algorithmic}
\usepackage{booktabs}
\usepackage{subfigure}
\usepackage{comment}
\usepackage{mathtools}
\usepackage{makecell}
\usepackage{amsmath}
\usepackage{amsfonts}
\usepackage{bm}

\usepackage{amsmath}

\usepackage{algorithmic}
\usepackage{times}
\usepackage{epsfig}
\usepackage{graphicx}
\usepackage{amsmath}
\usepackage{amssymb}
\usepackage{booktabs}
\usepackage{diagbox}
\usepackage{booktabs}

\usepackage{comment}
\usepackage{times}
\usepackage{epsfig}
\usepackage{graphicx}
\usepackage{amsmath}
\usepackage{arydshln}
\usepackage{amssymb}
\usepackage{multirow}
\usepackage[vlined,boxed,commentsnumbered,ruled,linesnumbered]{algorithm2e}
\usepackage{color}

\title{A Survey for Solving Mixed Integer Programming via Machine Learning}
\author{}
\date{December 2021}

\author{
Jiayi Zhang$^{1}$\and
Chang Liu$^{1}$\and
Junchi Yan$^1$\footnote{
Correspondence author is Junchi Yan.}\\
Xijun Li$^2$\and
Hui-Ling Zhen$^2$\and
Mingxuan Yuan$^2$
\affiliations
$^1$Department of CSE, and MoE Key Lab of Artificial Intelligence, 
Shanghai Jiao Tong University\\
$^2$Noah's Ark Lab, Huawei Ltd.\\
\emails  
\{zhangjiayirr,\ only-changer,\ yanjunchi\}@sjtu.edu.cn
\{xijun.li,zhenhuiling2,yuan.mingxuan\}@huawei.com
}

\begin{document}

\maketitle

\begin{abstract}
This paper surveys the trend of leveraging machine learning to solve mixed integer programming (MIP) problems. Theoretically, MIP is an NP-hard problem, and most of the combinatorial optimization (CO) problems can be formulated as the MIP. Like other CO problems, the human-designed heuristic algorithms for MIP rely on good initial solutions and cost a lot of computational resources. Therefore, we consider applying machine learning methods to solve MIP, since ML-enhanced approaches can provide the solution based on the typical patterns from the historical data. In this paper, we first introduce the formulation and preliminaries of MIP and several traditional algorithms to solve MIP. Then, we advocate further promoting the different integration of machine learning and MIP and introducing related learning-based methods, which can be classified into exact algorithms and heuristic algorithms. Finally, we propose the outlook for learning-based MIP solvers, direction towards more combinatorial optimization problems beyond MIP, and also the mutual embrace of traditional solvers and machine learning components. 

\end{abstract}

\section{Introduction}
Mixed integer programming problems (MIP) are significant parts of combinatorial optimization (CO) problems. Benefiting from the development of academic theory and commercial software, MIP has become a vital capability that powers a wide range of applications, including planning~\cite{ppmip,article2013}, scheduling~\cite{sscumip,mipf}, routing~\cite{article,7076321} and bin packing~\cite{GAJDA2022102559}. Moreover, \cite{paulus2021} manages to integrate integer programming solvers into neural network architectures as a differentiable layer capable of learning both the cost terms and the constraints. The resulting end-to-end trainable architectures are able to simultaneously extract features from raw data and learn a suitable set of constraints that specify any combinatorial problem. This architecture can learn to fit the right NP-hard problem needed to solve the task, which demonstrate the importance of MIP in the area of combinatorial optimization. 

When dealing with mathematical modeling problems in industrial applications mentioned above, decision variables and integer variables are sometimes inevitable, which are modeled as MIP models. For example: $x$ cars, $y$ packages. $x$, $y$ are integer variables here, and decimals are meaningless. One MIP problem may contain both integer and continuous variables. If the problem contains an objective function with no quadratic term, then the problem is termed a Mixed Integer Linear Programming (MILP). Formally, it is given as:
\begin{equation}
\begin{aligned}
&\min c^{\mathrm{T}} x \\
\text {s.t. }A x \geq b , & x \in \mathbb{R}^{n} \quad x_{j} \in \mathbb{Z}, \forall j \in I
\end{aligned}
\label{eq:mip}
\end{equation}

When there is at least a quadratic term in the objective, the problem terms to a Mixed Integer Quadratic Program (MIQP). If the model has any constraint containing a quadratic term, regardless of the objective function, the problem is termed as a Mixed Integer Quadratic Constrained Program (MIQCP). In this paper, we only focus on the basic MILP, and the MIP term denotes MILP from now on.

A lower bound for optimum of MILP is provided by solving the corresponding LP relaxation of the MILP. The LP relaxation of is obtained by omitted the integer requirements:
\begin{equation}
\begin{aligned}
&\min c^{\mathrm{T}} x \\
\text {s.t. }&A x \geq b , x \in \mathbb{R}^{n}
\end{aligned}
\end{equation}

Exactly solving the MILP is NP-hard, that is, exponential time solvable. While practitioners have more interested in getting solutions of good quality as quickly as possible than in finding a provably optimal solution. The research on how to obtain high-quality solutions in a limited time is of practical  great significance.

In this survey, we focus on ML-based algorithmic techniques to solve MIP. These basic and well-studied techniques can be divided into two main categories, \textit{the branch and bound based algorithms for exact solving} and \textit{the heuristic algorithms for approximately solving}. We will introduce one by one in the next sections, and discuss other interesting directions later. In modern MIP solvers, most of them are based on the branch-and-bound method, and the cutting plane technique is often 
integrated into the branch-and-bound method to increase the efficiency of branch-and-bound algorithm. These two techniques constitute the branch-and-cut framework. The common MIP solvers includes SCIP~\cite{BestuzhevaEtal2021ZR}, CPLEX~\cite{cplex2009v12} and GUROBI~\cite{llc2020gurobi}. Due to space limitations, this paper will not cover classical methods used in commercial solvers for solving MIP. 

There are emerging open-source libraries to facilitate the research at the intersection of machine learning and mixed integer programming.  MIPLearn\footnote{https://anl-ceeesa.github.io/MIPLearn} offers a customized library for learning-based solver configuration which supports both GUROBI and CPLEX. It can be framed as a borderline case of an MDP framework which is the focus of another emerging library Ecole~\cite{Ecole}. Ecole is designed to cooperate with the state-of-the-art open-source MIP solver SCIP where Ecole acts as an improvement to existing solvers. Besides, ORGym \cite{OR-gym20} and OpenGraphGym \cite{ZhengICCS20} are Gym-like libraries for learning heuristics for a collection of combinatorial optimization problems including MIP.

{\textbf{Difference to existing surveys.} There are excellent surveys in the area of machine learning for combinatorial optimization~\cite{BengioEJOR21}, and exactly learning for solving mixed integer programming~\cite{EJORSurvey21,TopSurvey17,huang2021branch}. Compared with the previous survey \cite{TopSurvey17} which only covers the topic of learning-supported node selection techniques due to the limited development five years ago. While the other ~\cite{EJORSurvey21} is focused on bi-level tailored approaches that exploit MIP techniques to solve bi-level optimization problems. In this paper, we aim to give a comprehensive and up-to-date review in the area of learning for solving MIP, especially seeing the rapid development in recent years. The recent paper \cite{huang2021branch} mainly focuses on the development in recent decades of the branch-and-bound algorithm instead of MIP, but it does not emphasize the importance of adapting machine learning in MIP.} 

\begin{table*}[htbp]
  \centering
  \renewcommand{\arraystretch}{1.3}
  \caption{Summary of methods that combine machine learning with B\&B. ``Selection" denotes which part involves learning.}
  \resizebox{\linewidth}{!}{
    \begin{tabular}{l|l|l|l|l|l}
    \toprule
    Method & Selection & Learning & Network & Representation & Remark \\
    \hline
    \cite{DBLP:journals/corr/abs-1906-01629} & Variable & Reinforcement & GCN   & Bipartite graph & Imitate strong branching \\
    \cite{DBLP:journals/corr/abs-2006-15212} & Variable & Supervised & GCN   & Bipartite graph  & Accelerate via dynamic embedding  \\
    \cite{sun2020improving} & Variable & Reinforcement &  PD policy & Subproblem set & Evolution strategy for training\\
    \cite{NIPS2014learn} & Node & Reinforcement & Standalone & Standalone & Imitate optimal oracle\\
    \cite{DBLP:journals/corr/abs-2007-03948} & Node & Supervised  & MLP   & Handcraft & Prune leaf \\
    \cite{10.5555/3015812.3015920} & Variable & Supervised & SVM   & Handcraft & Learning to rank \\
    \cite{shen2021learning} & Variable & Supervised  & GCN   & Bipartite graph & Combined with DFS \\
    \cite{huang2021learning} & Cutting & Supervised & MLP   & Handcraft & Large scale \\
    \cite{DBLP:journals/corr/abs-2002-05120} & Variable & Reinforcement & MLP   & Handcraft & Imitate strong branching \\
    \cite{pmlr-v119-tang20a} & Cutting & Reinforcement & Attention \& LSTM & Handcraft & Evolution strategy for training \\
    \cite{nair2021solving} & Variable & Reinforcement & GCN   & Bipartite graph & Imitate strong branching \\
    \cite{DBLP:journals/corr/abs-1906-09575} & Variable & Supervised
    & GCN & Tripartite graph & Extract connection information \\
    \cite{alvarez2014supervised}  & Variable & Supervised & ExtraTrees & Handcraft & Imitate strong branching \\
    \bottomrule
    \end{tabular}%
    
    }
  \label{tab1}%
\end{table*}%

\begin{figure}[t!] 
\centering 
\includegraphics[width=0.48\textwidth]{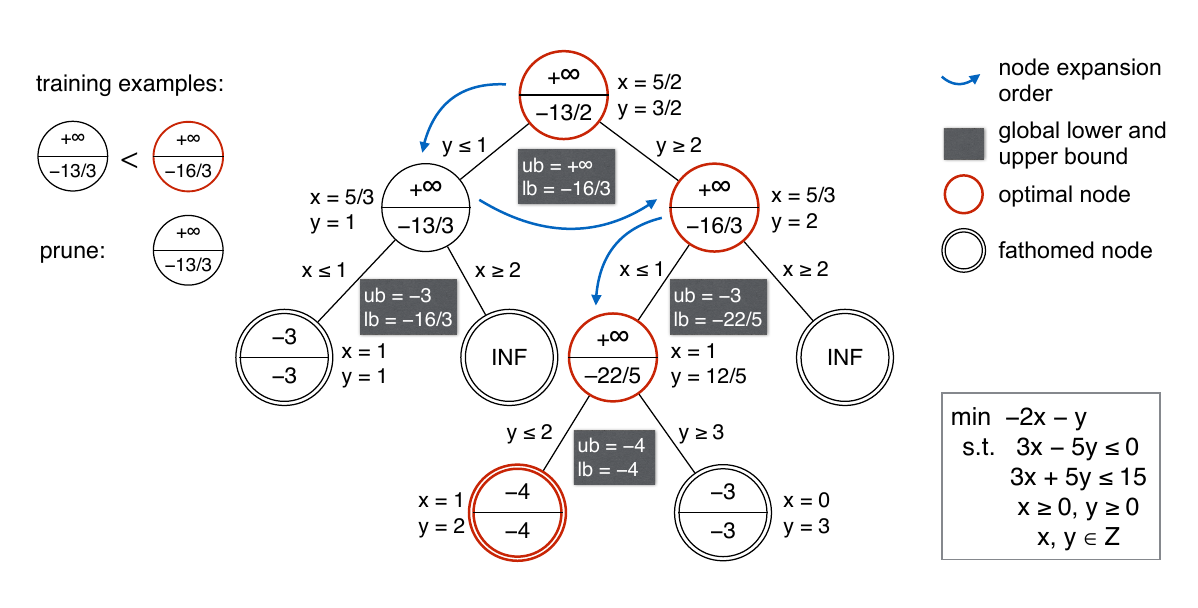}
\vspace{-20pt}
\caption{Use branch-and-bound to solve an integer linear programming minimization. (Credit to~\protect\cite{NIPS2014learn})} 
\vspace{-5pt}
\label{Fig.bandb} 
\end{figure}
\section{Branch-and-Bound}
In general, mixed integer linear programming problems can be solved using a linear-programming based branch-and-bound (B\&B) algorithm~\cite{Land2010}. Proposed half a century ago~\cite{Land1960AnAM}, B\&B is an exact algorithm that is proved to reach the optimal solution eventually. In this section, we first give an overview of the general B\&B algorithm, and then we divide the existing works into branching variable selection, node selection, cutting plane selection and discuss these methods combined with machine learning approaches in detail.

Branch-and-bound is a classic tree search method. By divide-and-conquer, it partitions the search space by branching on variables’ values and smartly uses bounds from problem relaxations to prune unpromising regions from the tree.

Define an optimization problem as $\mathcal{P}=(X, f)$, where $X$ (called the search space) is a set of valid solutions, and $f: X \rightarrow \mathbb{R}$ is the objective function. The goal is to find an optimal solution $x^{*} \in \arg \min _{x \in X} f(x)$. In order to solve $\mathcal{P}$, a search tree $T$ of subsets of the search space is build. Additionally, a feasible solution $\hat{x} \in X$ (called the incumbent solution) is stored globally. At each iteration, the algorithm selects a new subsets $S \subseteq X$ to explore from a list $L$ of unexplored subsets; if a solution $\hat{x}^{\prime} \in S$ (called a candidate incumbent) can be found with a better objective value than $\hat{x}$ (i.e., $f\left(\hat{x}^{\prime}\right)<f(\hat{x})$ ), the incumbent solution is updated. On the other hand, if it can be proven that no solution in $S$ has a better objective value than $\hat{x}($ i.e. $, \forall x \in S, f(x) \geq f(\hat{x}))$, the subset is pruned (or fathomed), and the subset is terminal. Otherwise, child subsets are generated by partitioning $S$ into an exhaustive (but not necessarily mutually exclusive) set of subset $S_{1}, S_{2}, \ldots, S_{r}$, which are then inserted into $T$. Once no unexplored subset remain, the best incumbent solution is returned; since the subset is only fathomed if they contain no solution better than $\hat{x}$, it must be the case that $\hat{x} \in \arg \min _{x \in X} f(x)$. Pseudocode for the generic B\&B procedure is given in Alg.~\ref{alg:bb}, and an example of solving MIP by B\&B~\cite{NIPS2014learn} is shown in Fig.~\ref{Fig.bandb}.

\begin{algorithm}[t]
\caption{\textbf{Branch-and-Bound $(X, f)$ }}
\label{alg:bb}
Set $L=\{X\}$ and\\
\While{$L \neq \emptyset$}
{
Select a subproblem $S$ from $L$ to explore\\
\If{a solution $\hat{x}^{\prime} \in\{x \in S \mid f(x)<f(\hat{x})\}$ can be found}
{Set $\hat{x}=\hat{x}^{\prime}$}
\If{$S$ cannot be pruned}
{
Partition $S$ into $S_{1}, S_{2}, \ldots, S_{r}$\\
Insert $S_{1}, S_{2}, \ldots, S_{r}$ into $L$ 
}
Remove $S$ from $L$
}
\KwOut{$\hat{x}$ }
\end{algorithm}

One critical reason why B\&B is difficult to formalize resides in its inherent exponential nature: millions of Branching Variable Selection (BVS) decisions could be needed to solve a MILP, and a single bad one could result in a doubled tree size and no improvement in the search. Such a complex and data-rich setting, paired with often a lack of formal understanding of the particular problem structure, makes B\&B an appealing ground for machine learning techniques, which has lately been thriving in discrete optimization. Employing a clever branching rule is critical for MIP solvers.

The majority of works focus on decision in the branch-and-bound algorithm~\cite{pmlr-v80-balcan18a,10.5555/3015812.3015920}. Two most crucial ones are branching variable selection and node selection. Recent works explore ``learning to branch", including learning for branching variable selection~\cite{10.5555/3015812.3015920,DBLP:journals/corr/abs-1907-04484,LIBERTO2016943,DBLP:journals/corr/abs-1906-01629,sun2020improving,DBLP:journals/corr/abs-2002-05120,DBLP:journals/corr/abs-2005-10026}, node selection~\cite{NIPS2014learn,DBLP:journals/corr/abs-1804-00846,DBLP:journals/corr/abs-1907-04484,DBLP:journals/corr/abs-2007-03948}, and cutting plane selection~\cite{huang2021learning,pmlr-v119-tang20a}. In particular, branching variable selection is the main 
stream of the presented articles, since the selection of the next variable is significant in B\&B and matters to the total cost of B\&B.

\subsection{Branching Variable Selection (BVS)}
\label{sec:bvs}
Branch variable selection determines which fractional variables (also known as candidates) to branch the current node into two child nodes. To indicate the quality of a candidate variable, a score of this variable is used to measure its effectiveness, and the candidate with the highest score is picked to branch on. The pseudocode of BVS is presented in Alg.~\ref{alg:bvs}, which illustrates the basic variable selection idea. 

\begin{algorithm}[t]
\caption{\textbf{Branching Variable Selection}}
\label{alg:bvs}
\KwIn{Subproblem of the current node $\mathcal{S}$ with its optimal LP solution $\hat{x} \notin X_{M I L P}$}
Define branching candidates set $C=\left\{i \in I \mid \hat{x}_{i} \notin \mathbb{Z}\right\}$ \\
\For{each candidate $i \in C$}
{
Calculate its score value $s_{i} \in \mathbb{R}$
}
\KwOut{A subscript $i=\arg \min _{i \in C} s_{i}$ of an integer variable with fractional value $\hat{x}_{i} \notin \mathbb{Z}$}
\end{algorithm}

There are various branching rules used to measures variables in BVS, including strong branching (SB) rule~\cite{10.5555/868329}, pseudo-cost~\cite{benichou1971experiments} and hybrid branching~\cite{achterberg2009hybrid}. Naturally, the idea of adopting an imitation learning~\cite{Ho2016GenerativeAI} strategy to learn a fast approximation of branching rules came into being. Imitation learning aims at using expert experience to conduct the learning of neural networks, which can acquire the neural networks that can reach the same or even better performance than the initial expert. \cite{alvarez2014supervised,alvarez2017machine,10.5555/3015812.3015920,DBLP:journals/corr/abs-1906-01629,DBLP:journals/corr/abs-2006-15212} learn branching policies by imitating the strong branching rule. This kind of approach adopts a high-quality but expensive heuristic scheme, which is the earliest and most widely studied method for machine learning. 

Specifically, \cite{alvarez2014supervised} is the first to use supervised learning to learn a strong branching model. By the observation of the expert, the neural networks can learn an approximated function of the variable score, and can further adapt it to select the suitable variable in B\&B. \cite{10.5555/3015812.3015920} extends the work of ~\cite{alvarez2014supervised} by designing a novel pipeline that can solve the MIP on the fly. In the first 500 branches, they use the traditional SB to make decisions while recording the problem features to train the neural networks, and the neural networks take control of the decision-maker from the 501st branch. Both of them imitate SB successfully, while the spending time is too high due to the calculation of a large number of features on each node.

\begin{figure}[t] 
\centering 
\includegraphics[width=0.5\textwidth]{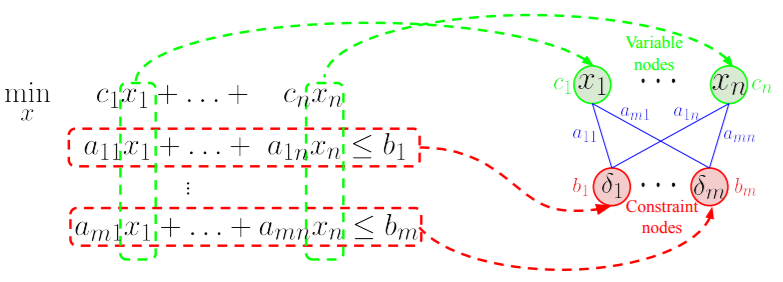}
\vspace{-20pt}
\caption{Transform an MIP instance to a bipartite graph. The variables (in green) correspond to one side and the constraints (in red) refers to the other. (Credit to~\protect\cite{nair2021solving})} 
\label{Fig.bp} 
\end{figure}

\begin{figure}[tb!] 
\centering 
\includegraphics[width=0.48\textwidth]{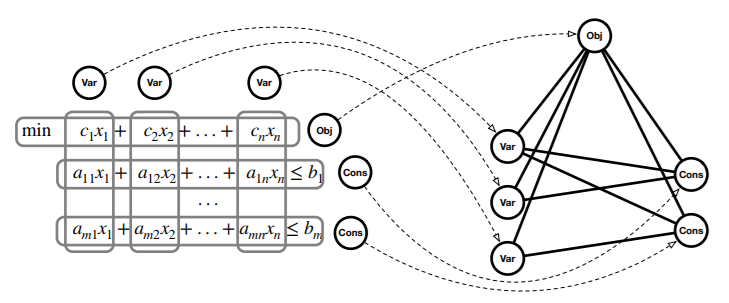}
\vspace{-20pt}
\caption{From MIP to tripartite graph. Three types of vertices represent objective functions, variables, and constraints, respectively. (Credit to~\protect\cite{DBLP:journals/corr/abs-1906-09575})} 
\vspace{-5pt}
\label{Fig.tp} 
\end{figure}

To overcome complex feature calculation,~\cite{DBLP:journals/corr/abs-1906-01629} encodes MIP into a graph-convolutional network (GCN)~\cite{Kipf2017SemiSupervisedCW}, which can extract the features on the graph efficiently by various message passing approaches. \cite{nair2021solving} encodes MIP to the graph-convolutional network (GCN) as a bipartite graph and computes an initial feasible solution (Neural Diving), then trains a GCN to imitate ADMM-based policy for branching (Neural Branching). It uses RL to learn a policy that un-assigns one variable at a time, interleaved with solving a sub-MIP every $\eta$ steps to compute a new solution. The authors formulate MIP as a natural bipartite graph representing variable-constraint relationships, which is illustrated in Fig.~\ref{Fig.bp}. Each MIP has two sets of nodes, one representing variables and another representing constraints. And an edge between a variable $i$ and a constraint $j$ means variable $i$ appears in constraint $j$. 
A lot of subsequent work has referenced the bipartite graph model proposed by ~\cite{DBLP:journals/corr/abs-1906-01629}. Inspired by the utilization of the GCN model, ~\cite{DBLP:journals/corr/abs-2006-15212} proposes a hybrid architecture that uses a GNN model at only the root node of the B\&B tree and a weak but fast predictor at the remaining nodes, such as a simple Multi-Layer Perception (MLP). The model combines the superior representation framework of ~\cite{DBLP:journals/corr/abs-1906-01629} with the computationally cheaper framework of~\cite{10.5555/3015812.3015920} to realize a time-accuracy trade-off in branching. Besides, \cite{DBLP:journals/corr/abs-1906-09575} generates a tripartite graph from its MIP formulation, as illustrated in Fig.~\ref{Fig.tp}. They train a GCN for variable solution prediction on the collected features, labels, and tripartite graphs. In~\cite{DBLP:journals/corr/abs-2002-05120}, the authors claim that information contained in the global branch-and-bound tree state is an important factor in variable selection, and they use a novel neural network design that incorporates the branch-and-bound tree state directly. Furthermore, they are one of the few techniques on heterogeneous instances.

Different from learning the output of a computationally expensive heuristic used in B\&B, \cite{DBLP:journals/corr/abs-2005-10026} applies reinforcement learning (RL) for BVS from scratch, free from any heuristic.
The core idea of RL is to learn from the interactions between the agent and the environment, and here the MIP instance is the environment and the agent is try to solve it. Experience shows that the RL approach sometimes works better than the imitation learning we mentioned before. In \cite{sun2020improving}, the authors argue that strong branching is not a good expert to imitate and utilize RL by modeling the variable selection process as a Markov Decision Process (MDP). They further design a policy net based on primal-dual iteration over reduced LP relaxation, which utilizes the power of evolution strategy to update the neural networks.

\begin{table*}[tb!]
  \centering
  \renewcommand{\arraystretch}{1.3}
  \caption{A brief summary of methods that combine machine learning with heuristic algorithms. The ``Focus" column denotes the combined heuristic algorithm. LNS means large neighborhood search and FP means feasibility pump.}

  \resizebox{\linewidth}{!}{
    \begin{tabular}{l|l|l|l|l|l}
    \toprule
    Method & Focus & Learning & Network & Representation & Remark \\
    \hline
    \cite{song2020general} & LNS & Reinforcement & MLP   & Handcraft & Combined with GUROBI \\
    \cite{sonnerat2021learning} & LNS   & Reinforcement & GCN   & Bipartite graph & Use imitation learning \\
    \cite{wu2021learning} & LNS   & Reinforcement & GCN   & Bipartite graph & Outperforms SCIP \\
    \cite{Liu2021LearningTS} & LNS   & Supervised \& Reinforcement  & GNN   & Bipartite graph & Adaptive neighborhood size  \\
    \cite{Qi2021ReinforcementLF} & FP    & Reinforcement & MLP \& CNN & Parameter matrix & Combine [A,b] as matrix \\
    \cite{ding2020accelerating} & Pick & Supervised& GCN   & Tripartite graph & Predict solution value for variables \\
    \cite{grover2018best} & Pick & Reinforcement  & Standalone & Handcraft & Pick heuristics in CPLEX \\
    \cite{xavier2021learning} & Pick & Supervised & KNN   & Handcraft & As a warm start \\
    \bottomrule
    \end{tabular}%
    }
  \label{tab2}%
\end{table*}%

\subsection{Node Selection}
As mentioned above, the branch-and-bound algorithm recursively divides the feasible set of a problem into disjoint subsets, organized in a tree structure, where each node represents a subproblem that searches only the subset at that node. The main steps of the node selection algorithm are given in Alg.~\ref{alg:ns}. If computing bounds on a subproblem does not rule out the possibility that its subset contains the optimal solution, the subset can be further partitioned (``branched”) as needed. Else if the lower bound of possible solutions of a node is larger than the known upper bound, the node can be pruned. A key question in B\&B is how to prioritize which nodes to consider. An effective node priority decision strategy guides the tree search to promising areas and improves the chance of quickly finding a good incumbent solution, which can be used to decide whether to discard or expand other nodes. Thus, learning the appropriate node selection policy of a B\&B tree is worthy of being investigated. More precisely, a policy is a function that maps a state to an action, which in this case is the next node to be selected.

\begin{algorithm}[t!]
\caption{\textbf{Node Selection}}
\label{alg:ns}
\KwIn{Node list $L = \{P\}$ and  upper bound $\bar{J}$}
Assigns a score to each active node \\
Pop a node $P^{i}$ from $L$ \\
Solve the LP relaxation \\
get solution$\left(x^{i}, y^{i}\right)$and lower bound $J^{i}$ \\
\If{$J^{i} \geq \bar{J}$}
{
\quad Prune the node $P^{i}$
}
\ElseIf{$y^{i} \in \mathbb{Z}^{p}$}
{
Improve upper bound $\bar{J} = J^{i}$
}
\Else{Branch node $P^{i}$ and push child nodes to list $L$}
\end{algorithm}

There is less work in node selection of B\&B than in BVS. \cite{NIPS2014learn} uses imitation learning to train a node selection and a node pruning policy to speed up the tree search in the B\&B process.
Following the above work,~\cite{DBLP:journals/corr/abs-2007-03948} obtains a node selector by imitation learning. The difference is that ~\cite{DBLP:journals/corr/abs-2007-03948} learns only to choose a node’s children it should select. This encourages finding solutions quickly, as opposed to learning a BFS-like method.

\begin{figure}[t!] 
\centering 
\includegraphics[width=0.45\textwidth]{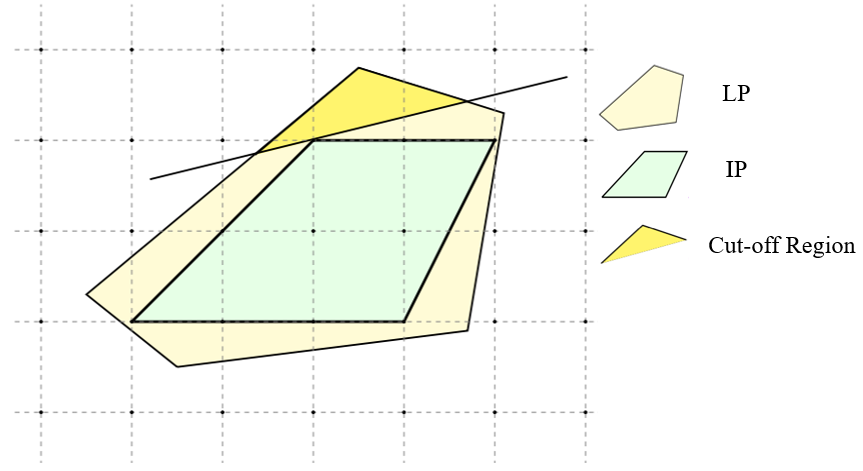}
\vspace{-10pt}
\caption{Cutting planes for MIP.} 
\vspace{-5pt}
\label{Fig.cp} 
\end{figure}

\subsection{Cutting Plane}
It is known that every MIP can be relaxed to a linear programming (LP) by dropping the integer constraints, and there are many traditional efficient algorithms for solving LP such as Simplex~\cite{dantzig1990origins}, Interior Point Method (IPM)~\cite{roos2005interior}, etc. It means that we can relax MIP to LP and solve the incident LP instead at an acceptable cost, which leads to the cutting plane algorithm. As Fig.~\ref{Fig.cp} shows, cutting plane methods iteratively add cuts to the relaxed LP, which are linear constraints that can tighten the LP relaxation by eliminating some part of the feasible region while preserving the LP optimal solution. Suppose that we add the cut set $C^{'} = \left\{\alpha_{i}^{T} x \geq \beta_{i}\right\}_{i=1}^{|C^{'}|} $ to the original formulation in Eq.~(\ref{eq:mip}). Then, the optimization formulation of MIP becomes:
\begin{equation}
\begin{aligned}
&\min c^{\mathrm{T}} x \\
\text {s.t. }A x \geq b , x \in \mathbb{R}^{n} \quad
&\alpha^{T} x \geq \beta, \quad x_{j} \in \mathbb{Z}, \forall j \in I
\end{aligned}
\label{eq:mip_cp}
\end{equation}

As shown by the new formulation, cuts serve as the purpose of reducing the LP solution space, which might lead to a smaller tree in the branch-and-cut algorithm so that the number of nodes to be searched is significantly reduced. As mentioned before, the cutting plane can be combined with the branch and bound algorithm, which constitutes the branch-and-cut framework. Branch-and-cut is known as one of the most commonly used algorithms in modern solvers.

Due to the importance of the cutting plane in solving MIP, many researchers try to utilize machine learning technologies to improve the traditional cutting plane algorithm. Unlike B\&B, there are relatively few well-designated heuristics in the cutting plane algorithm, which denotes imitation learning or supervised learning cannot be directly applied here. Therefore, researchers begin to think of reinforcement learning. \cite{pmlr-v119-tang20a} introduces a MDP formulation for the problem of sequentially selecting cutting planes for MIP, and training a reinforcement learning (RL) agent using evolutionary strategies. This work shows the ability of RL to improve the cutting plane algorithm and potentially opens a new research topic. The following work \cite{huang2021learning} proposes a cut ranking method for cut selection in the settings of multiple instance learning. It uses neural networks to give scores to different candidate cuts for the next step.

\section{Heuristic Algorithms}
Though MIP can be solved via B\&B exactly, it is time and resource-consuming due to its NP-hard nature. In many cases especially large-scale problem instances, B\&B becomes intractable. Thus, heuristic-based methods are considered instead. Heuristic algorithms aim to solve MIP approximately, by the integration of greedy approach and searching. The common heuristic algorithms for MIP are large neighborhood search and the feasibility pump. Besides, some works aim at better utilizing existing MIP solvers by machine learning. We will discuss these approaches in detail. 


\subsection{Large Neighborhood Search (LNS)}
Large neighborhood search (LNS) is a powerful heuristic for MIP. Given the problem instance and the initial feasible solution, LNS searches for better candidate solutions among pre-defined neighborhoods of current solution in each iteration. The iteration continues until the search budget (for example the computing time) is used up. Due to the nature of LNS, it is important to prevent the search from falling into a poor local optimum. In general, the size of the neighborhood grows exponentially as the size of the input problem increase. Therefore, it is necessary to optimize the LNS algorithm by the learning techniques to improve its efficiency.

There are two critical choices to determine the effectiveness of LNS: 1) initial solution and 2) search neighborhood at each iteration. \cite{song2020general} learns a neighborhood selection policy using imitation learning and reinforcement learning (RL). It uses a random neighborhood selection policy to generate training data for imitation learning. 
Following the neural diving idea proposed by~\cite{nair2021solving} mentioned in Section~\ref{sec:bvs}, \cite{sonnerat2021learning} adapts neural diving to obtain the initial solution and for selecting the search neighborhood at each LNS step. Some researchers~\cite{Liu2021LearningTS} conduct analysis to LNS, and it turns out the size of neighbors is important, and the most suitable neighbor size varies over iterations. Therefore, they propose to use machine learning to automatically find the suitable neighbor size. \cite{wu2021learning} combines RL with LNS, where the action of RL is to select the variable to be replaced. They also propose a novel feature extractor for variables and constraints in MIP. 

LNS aims to continuously improve a solution, which is a common idea in solving CO problems. Therefore, we will discuss some approaches in the CO field that shares similar ideas with LNS, which we hope could inspire adaptation of these methods to solve MIP. \cite{Chen2019LearningTP} proposes a framework called local rewrite, which tries to improve a given solution by selecting a part of the solution and modifying it. In their paper, the local rewrite framework is proved to be a powerful method for the vehicle routing problem and computing resource allocation. The ECO-DQN~\cite{barrett2020exploratory} framework re-designs the action space of the reinforcement learning agent, which allows revoking the previous action. In other words, the action of the agent is revocable in the ECO-DQN framework. There are many other works~\cite{Lu2020ALI,Fu2021GeneralizeAS} following the idea of local rewrite and ECO-DQN in combinatorial optimization.


\begin{algorithm}[t]

\caption{\textbf{Feasibility Pump}}
\label{alg:lns}
\KwIn{$x^{0} \leftarrow argmin \ c^{T}x;\ \overline{x}^{0} \leftarrow [x^{0}];\ i = 0$}
\While{$\overline{x}^{i}$ is not feasiable}
{
$x^{i + 1} \leftarrow argmin \ ||x - \overline{x}^{i}||$\\
$\overline{x}^{i + 1} \leftarrow [x^{i + 1}]$ \\
\If{$\overline{x}^{i + 1} ==\ \overline{x}^{i}$}
{
random perturbation of $\overline{x}^{i}_{j}, \forall j \in I$
}
\Else{$k \leftarrow k + 1$}
}
\KwOut{$\overline{x}^{i}$}
\end{algorithm}

\subsection{Feasibility Pump}
Feasibility pump (FP) is a heuristic algorithm that runs the following steps: 1) find the rounded optimal continuous relaxation solution of the MIP (degenerating to LP); 2) search for the nearest solution in the relaxed feasible region; 3) perturb and round the new solution found at each step until the solution is feasible. If the limit of the maximum number of steps is reached, the algorithm will halt and return the current feasible solution as the output. The basic steps of the FP algorithm are shown in Algorithm~\ref{alg:lns}.

Though FP is a relatively powerful heuristic, the efficiency and cost-effectiveness are not satisfied. Therefore, \cite{qi2021smart,Qi2021ReinforcementLF} utilize RL to improve FP. The entry point of RL is to find the next non-integer solution in step 2). Instead of choosing the nearest solution, the authors let the RL agent choose the next solution as its action. Besides, two methods of MLP and CNN are designed on the representation of the state space, where the CNN is to treat the parameters [A, b] in MIP as an image and process it. Fig.~\ref{Fig.cp} shows the FP algorithm and the smart FP proposed by~\cite{qi2021smart}.

\begin{figure}[t!] 
\centering 
\includegraphics[width=0.4\textwidth]{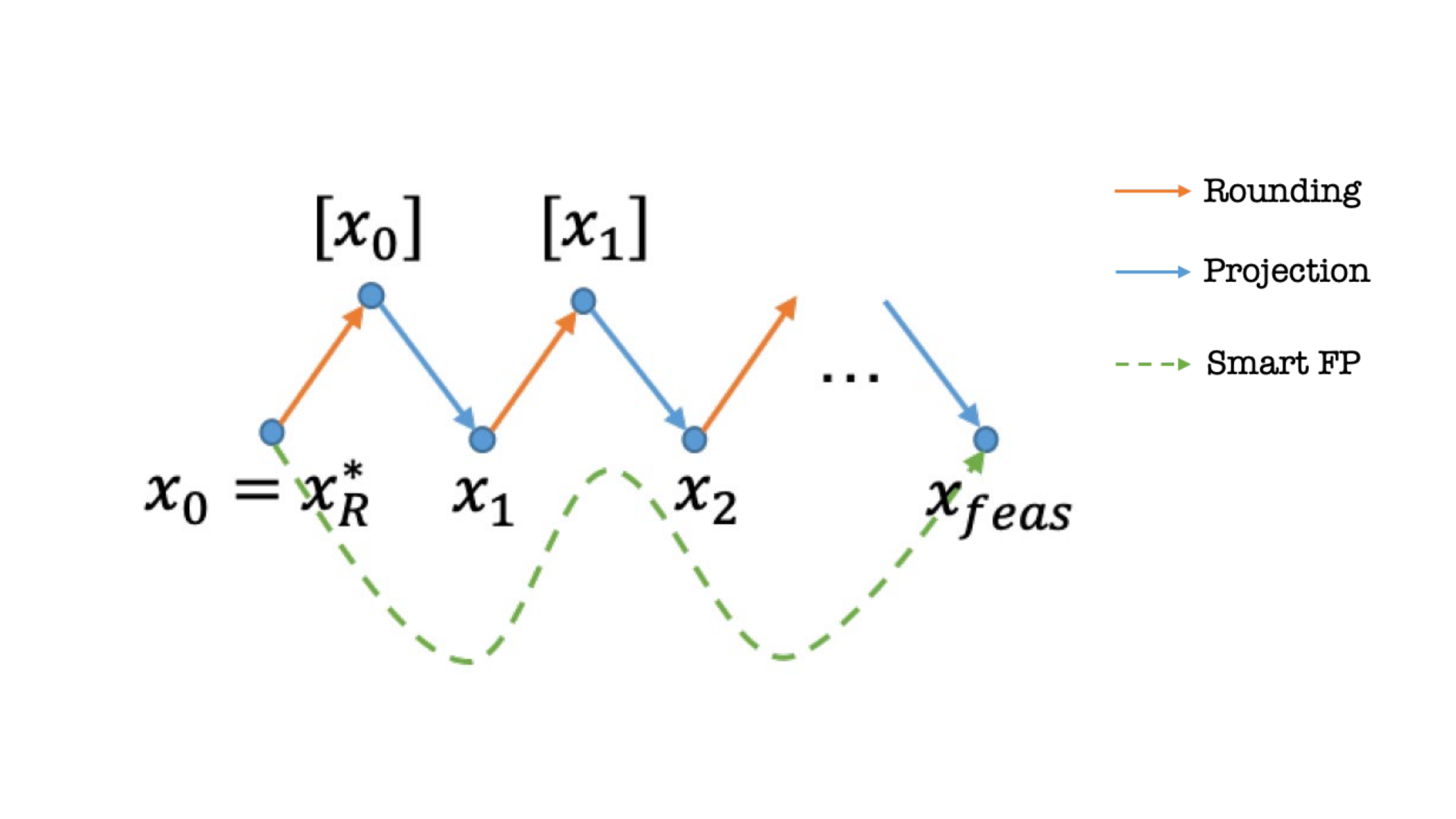}
\vspace{-30pt}
\caption{The feasibility pump. (Credit to~\protect\cite{qi2021smart})} 
\vspace{-10pt}
\label{Fig.fp} 
\end{figure}

\subsection{Predict and Pick}
Instead of solving the MIP directly, some works aim to predict and pick from the existing MIP solvers or methods. It is reasonable to solve MIP based on existing works since existing MIP solvers have been improved for decades. Some researchers believe that better utilization of existing MIP methods is a valuable research topic. \cite{ding2020accelerating} predicts a solution value for each variable based on historical data and decides whether to use heuristic algorithms or the exact branching approach to solve it. The authors try to make the exact branching approach focus on the hard case while leaving the easy case to the heuristic algorithm. By adopting reinforcement learning, \cite{grover2018best} proposes to select the predefined heuristics in CPLEX by the features of the given instance, which is an online learning framework. \cite{xavier2021learning} focuses on solving the MIP instances in a data-driven manner. Combined with existing MIP solvers, they train a KNN to predict redundant constraints, good initial feasible solutions, and affine subspaces where the optimal solution is likely to lie, which can lead to a significant reduction in the problem size of MIP. They conduct experiments on the electric grid unit commitment problem, which is an application of MIP. Besides, the early stopping technique is widely adopted in the hyperparameter optimization~\cite{makarova2021overfitting}, which uses machine learning to predict when to stop the searching process without much loss of quality. It greatly improves the efficiency and saves computing resources. The early stopping technique might be integrated into the MIP solvers. Specifically, one can use machine learning techniques to predict when to early stop the B\&B process meanwhile the incumbent primal solution is still acceptable.

\section{Conclusion and Outlook}
In this survey, a study of the state-of-the-art machine learning approach for solving MIP is presented, to summarize existing work and serve as a starting point for future research in these areas. We find that the integration of machine learning techniques and traditional operational research algorithms is a raising topic in the research field, including combination with the exact B\&B algorithms and heuristic algorithms.

Since MIP is an NP-hard problem, it is very difficult to obtain an exact solution. Leveraging machine learning techniques to obtain an acceptable solution within limited computing resources is welcomed and reasonable in practical applications. We can design a model, by instructing learning models to imitate heuristic algorithms, and make some decisions in the B\&B algorithm or make adjustments to the initial solution. It can be seen that with the development of machine learning, especially for deep and reinforcement learning, more and more models are used in MIP solving, continuously improving the efficiency and solution quality.

\bibliographystyle{named}

\bibliography{mip}

\end{document}